\documentclass{l4dc_style/l4dc2024_arxiv}
\usepackage{times}

\usepackage[utf8]{inputenc}
\usepackage{natbib}

\usepackage{amsmath}

\usepackage{amsfonts}       
\usepackage{amssymb}
\usepackage{mathtools}
\usepackage{wrapfig}
\usepackage{booktabs}
\usepackage{caption}
\usepackage{hyperref}
\usepackage{ifthen}
\usepackage{fancyhdr}

\newboolean{arxiv}
\setboolean{arxiv}{true}

\author{%
 \Name{Philipp Pilar} \Email{philipp.pilar@it.uu.se}\\
 \addr Department of Information Technology, Uppsala University, Sweden
 \AND
 \Name{Niklas Wahlström} \Email{niklas.wahlstrom@it.uu.se}\\
 \addr Department of Information Technology, Uppsala University, Sweden
}

\title[Physics-informed Neural Networks with Unknown Measurement Noise]{Physics-informed Neural Networks with \\ Unknown Measurement Noise}

\begin{document}

\maketitle
\begin{abstract}
    Physics-informed neural networks (PINNs) constitute a flexible approach to both finding solutions and identifying parameters of partial differential equations.
    Most works on the topic assume noiseless data, or data contaminated with weak Gaussian noise.
    We show that the standard PINN framework breaks down in case of non-Gaussian noise.
    We give a way of resolving this fundamental issue and we propose to jointly train an energy-based model (EBM) to learn the correct noise distribution.
    We illustrate the improved performance of our approach using multiple examples.
    
\end{abstract}

\section{Introduction}

While the idea of using neural networks to solve partial differential equations (PDEs) dates back to the work of \citet{Lagaris1998}, the field has received renewed attention \citep{Cuomo2022, Karniadakis2021, Markidis2021, Blechschmidt2021} due to the seminal work of \citet{Raissi2019}, in which they introduced physics-informed neural networks (PINNs).
Both the forward problem, where the solution to a PDE is learned given the boundary conditions, as well as the inverse problem, where parameters of the PDE are to be inferred from measurements, can be solved with the PINN approach.

A multitude of applications for PINNs in science have already been considered:
\citet{Cai2021a} review the application of PINNs to fluid mechanics, \citet{Cai2021b} review their application to heat transfer problems, \citet{Yazdani2020} utilize PINNs to infer parameters for systems of differential equations in systems biology,
\citet{Mao2020} investigate the applicability of PINNs to high-speed flows, and
\citet{Costabal2020} utilize PINNs to take into account wave propagation dynamics in cardiac activation mapping.

The main advantage of PINNs over traditional solvers lies in their flexibility, especially when considering the inverse problem \citep{Karniadakis2021}:
as neural networks, they have the capacity for universal function approximation, they are mesh-free, and they can directly be applied to very different kinds of PDEs, without the need to use a custom solver.
When dealing with the inverse problem, parameters are learned from data and the question as to the effect of noisy data on the quality of the estimates arises naturally.
However, most of the existing work on PINNs assumes either noiseless data, or data contaminated with weak Gaussian noise.
While some research has been done on the effects of noisy data in PINN training (see Section \ref{sec:related_work}), they still consider either Gaussian noise or are focused on uncertainty quantification.

\begin{figure*}[!t]
\centering
	\includegraphics[width=\linewidth]{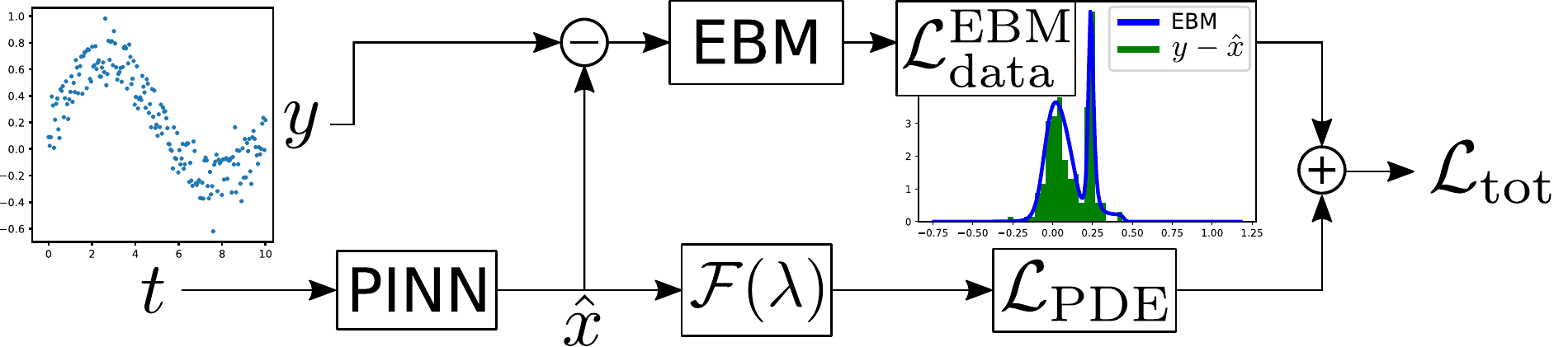}
	\caption{
	The PINN-EBM approach: 
	the inputs $t$ are passed through the PINN to obtain the PINN predictions $\hat x$.
	The differential operator $\mathcal{F}(\lambda)$ is then applied to $\hat x$ to obtain the PDE residuals, which are subsequently utilized to calculate the PDE loss $\mathcal{L}_{\rm PDE}$.
	At the same time, the residuals between the noisy measurements $y$ and $\hat x$ are formed, serving as noise estimates.
	The EBM is trained on these estimates in order to learn the noise PDF, which can in turn be utilized to compute the likelihood of the measurements serving as data loss $\mathcal{L}_{\rm data}^{\rm EBM}$.
	Finally, the two loss terms are combined to form the total loss $\mathcal{L}_{\rm tot}$.
	Both PINN and EBM, as well as the PDE parameters $\lambda$, can be trained by backpropagating~$\mathcal{L}_{\rm tot}$. \phantom{to not mess up formatting}
	}
	\label{fig:overview}
\end{figure*}

In this work, we consider the inverse problem for the case of measurements contaminated with homogeneous non-Gaussian noise of unknown form.
The least-squares loss, which is commonly employed as data loss in PINNs, is known to perform poorly in this case 
\citep{Constable1988, Akkaya2008}.
We give a way of mitigating this issue by suitably modeling the noise distribution.
A high-level illustration of our method is given in Fig. \ref{fig:overview}:
we employ an energy-based model (EBM) to learn the noise probability density function (PDF) jointly with the PINN.
This PDF is utilized to estimate the likelihood of the measurements under our model, which in turn serves as data loss.

Non-Gaussian measurement noise of unknown form can appear in a variety of applications \citep{johnson1991existence}.
Measurements of geomagnetic fields may exhibit asymmetric, long-tailed noise \citep{Constable1988},
data in astrophysics or seismology may be contaminated with impulsive noise \citep{weng2005nonlinear},
and not accounted for systematic errors in the measurement procedure may give rise to bias in the noise \citep{barlow2002systematic}.
For example, the noise distribution considered in Fig. \ref{fig:overview} can be interpreted as resulting from the case where measurements from multiple sensors have been merged into one dataset.
One of the sensors produces biased measurements, giving rise to the second peak away from zero.
Then the PINN-EBM allows for solving the PDE correctly together with the simultaneous detection of previously unrecognized systematic errors in the data.

\section{Related Work} \label{sec:related_work}

Prior research has been done on PINNs in case of noisy measurements, although typically only Gaussian noise is considered.
In \citet{Yang2021}, the framework of Bayesian neural networks \citep{Goan2020} is combined with the PINN framework, in order to obtain uncertainty estimates when training PINNs on noisy data.
\cite{psaros2023uncertainty} discuss different methods of uncertainty quantification and \cite{zou2023uncertainty} consider the case where both inputs and outputs are noisy.
\citet{bajaj2023recipes} introduce the GP-smoothed PINN, where a Gaussian process (GP) \citep{Rasmussen2006} is utilized to ameliorate noisy initial value data and make the PINN training more robust in this situation.
In \citet{Chen2021}, the PINN-SR method is introduced which can be employed to determine governing equations from scarce and noisy data.
In contrast to these works, we give a way of taking into account unknown, non-Gaussian noise in the PINN and provide a training procedure for this case.

While EBMs are commonly employed for classification \citep{Lecun2006} and image generation \citep{Du2019}, they have also been successfully applied to regression problems;
applications include object detection \citep{Gustafsson2020b} and visual tracking \citep{Danelljan2020}.
In \citet{Gustafsson2020a}, different methods for training EBMs for regression are discussed.
In our work we also consider regression tasks and to the best of our knowledge, our paper is the first to combine EBMs and PINNs.
The standard EBM approach to regression would not take the physical knowledge in form of the differential equation into account.

\section{Background} \label{sec:background}

In this section we give a brief introduction to the two distinct methods which we combine in our work: physics-informed neural networks and energy-based models.

\subsection{Physics-informed Neural Networks (PINNs)} \label{sec:PINN}
The PINN-framework \citep{Raissi2019} can be used as an alternate approach to solving PDEs, other than the standard, mesh-based solvers.
The objective is to numerically determine the solution to a differential equation $\mathcal{F}x(t) = 0$, where  $\mathcal{F}$ denotes the differential operator defining the PDE, $x(t)$ the solution to the differential equation, and $t$ the input; both $x$ and $t$ can be multidimensional.
In the PINN approach, we now employ a neural network to parameterize the numerical solution $\hat x(t) = \hat x(t| \theta_{\rm PINN})$ of the PDE, where $\theta_{\rm PINN}$ denotes the weights of the PINN which are to be optimized.

In this paper, we consider the inverse problem: 
we are given a dataset $\mathcal{D}_d = \{t_d^i, y_d^i\}_{i=1}^{N_d}$ of $N_d$ (noisy) measurements $y_d^i = x(t_d^i) + \epsilon$ of the PINN solution in addition to the parametric form of the differential operator $\mathcal{F}(\lambda)$, which may contain unknown parameters $\lambda$.
Furthermore, we choose another set of $N_c$ so-called collocation points $\mathcal{D}_c = \{t_c^i\}_{i=1}^{N_c}$. They can lie at arbitrary points in the input domain of the PDE and are used to take into account the PDE constraint. As long as collocation points are placed in the areas of interest, they can enable the PINN to extrapolate also to areas without measurements.

When training the PINN, two losses enter into the loss function:
the data loss, $\mathcal{L}_{\rm data}$, evaluating the fit of the PINN prediction with the data, and the PDE loss, $\mathcal{L}_{\rm PDE}$, a measure of the fulfillment of the PDE by the PINN solution.
In the standard PINN, the least-squares loss is used as data loss,
\begin{equation} \label{eq:data_loss}
    \mathcal{L}_{\rm data}(\hat x, \{t_d, y_d\}_{\rm mb}) = \frac{1}{N_d'} \sum_{i=1}^{N_d'} (\hat x(t_d^i) - y_d^i)^2,
\end{equation}
and the squares of the PDE residuals, $f(t) = \mathcal{F}(\lambda)\hat x(t) \overset{!}{=} 0$, at the collocation points serve as PDE loss,
\begin{equation} \label{eq:pde_loss}
    \mathcal{L}_{\rm PDE}(\mathcal{F}, \hat x, \{t_c\}_{\rm mb}) = \frac{1}{N_c'} \sum_{i=1}^{N_c'} f(t_c^i)^2.
\end{equation}
Here, $N_d'$ and $N_c'$ denote the number of data points $t_d$ and collocation points $t_c$, respectively, in the current mini-batch (mb).
The PINN is then trained by minimizing the total loss $\mathcal{L}_{\rm tot} = \mathcal{L}_{\rm data} + \omega \mathcal{L}_{\rm PDE}$ with respect to the parameters $\theta_{\rm PINN}$ and $\lambda$.
Unless stated otherwise, the weighting factor $\omega = 1$.
Utilizing this loss function, the PINN is optimized via some variation of gradient descent.

\subsection{Energy-based Models (EBMs)} \label{sec:EBM}

EBMs constitute a powerful method of learning probability densities from data \citep{Lecun2006}.
They are frequently employed for image generation and modeling \citep{Du2019, Nijkamp2020}, and have successfully been applied to the domain of regression (see Section \ref{sec:related_work}).
The EBM can retain all of the flexibility of a neural network, via the following parametrization:
\begin{equation} \label{eq:EBM_likelihood}
    p(y|t, \hat h) = \frac{e^{\hat h(t,y)}}{Z(t, \hat h)}, \qquad Z(t, \hat h) = \int e^{\hat h(t,\tilde y)} d\tilde y,
\end{equation}
where $\hat h(t,y) = \hat h (t, y| \theta_{\rm EBM})$ is the (scalar) output of the neural network with weights $\theta_{\rm EBM}$.

This expressive capacity, however, comes at the cost of a more complicated training procedure, since the partition function $Z(t, \hat h)$ will typically be analytically intractable.
In case of a high-dimensional $y$, it will also be expensive or infeasible numerically;
Monte Carlo methods are often employed to approximate the intractable integral in \eqref{eq:EBM_likelihood}.
Various ways of training EBMs for regression are given in \citet{Gustafsson2020a}.

In this paper, we will choose the approach of directly minimizing the negative log-likelihood (NLL), which can be written as
\begin{equation} \label{eq:NLL}
    \text{NLL}(\{t_d, y_d\}_{\rm mb}, \hat h) = - \log \left(\prod_{i=1}^{N_d'} p(y_d^i|t_d^i, \hat h) \right) =  \sum_{i=1}^{N_d'} \log{Z(t_d^i,\hat h)} -  \hat h(t_d^i, y_d^i).
\end{equation}
In our case, the evaluation of the partition function to high accuracy remains tractable by utilizing numerical integration, since the noise distribution is one-dimensional.
Higher-dimensional outputs are still possible, as long as the noise is uncorrelated.
Also, note that this does not constitute a restriction on the dimensionality of the inputs $t$ of the PDE.

\section{Problem Formulation} \label{sec:problem}

We consider the following setup, consisting of the differential equation and a measurement equation:
\begin{equation} \label{eq:pde}
    \mathcal{F}(\lambda)x(t) = 0, \quad \quad y(t) = x(t) + \epsilon, 
\end{equation}
where the parametric form of the differential operator $\mathcal{F}(\lambda)$ is given, as well as $N_d$ measurements $y(t)$ of the corresponding solution $x(t)$ contaminated with homogeneous measurement noise $\epsilon$. 
In this work, we aim to combine the PINN and the EBM framework in order to solve the inverse problem in case of measurements contaminated with non-Gaussian and non-zero mean noise.

\subsection{PINNs in Case of Non-zero Mean Noise} \label{sec:PINN_nonzero}
To see why noise with non-zero mean is problematic in the standard PINN framework, consider the loss function $\mathcal{L}_{\rm tot} = \mathcal{L}_{\rm data} + \omega \mathcal{L}_{\rm PDE}$ (compare Section \ref{sec:PINN}):
\begin{equation} \label{eq:lossges}
    \mathcal{L}_{\rm tot} =  \frac{1}{N_d'} \sum_{i=1}^{N_d'} \left(\hat x(t_d^i) - y_d^i \right)^2  +  \frac{\omega}{N_c'} \sum_{i=1}^{N_c'} f(t_c^i)^2 .
\end{equation}
A major challenge when training PINNs, even in case of Gaussian noise, lies in the fact that data and PDE loss counteract each other, since the first term would encourage overfitting towards each data point, whereas the second term would push the solution towards fulfilling the PDE.

Let us first consider the case of a zero-mean noise distribution in the limit of infinite data generated according to \eqref{eq:pde}, and assume that the PINN has sufficient capacity to represent the solution accurately.
Then both loss terms would be minimal for $\hat x(t) = x(t)$:
in the limit, the data loss would become $\underset{N_d \rightarrow \infty}{\lim}\mathcal{L}_{\rm data} = \mathbb{E}[(\hat x - x - \epsilon)^2] = \mathbb{E}[(\hat x - x)^2] + \text{Var}_{\epsilon}$, and hence $\hat x = x$.
The PDE loss would then obviously also be minimal (more precisely zero) for $\hat x = x$, due to \eqref{eq:pde}.

For noise with non-zero mean, however, the problem of mismatch between the two loss terms is not simply an effect of finite data, which could be treated via an adequate regularization procedure, but has deeper roots.
Considering again the limit of infinite data, we have $\underset{N_d \rightarrow \infty}{\lim}\mathcal{L}_{\rm data} = \mathbb{E}[(\hat x - x - \epsilon)^2] = \mathbb{E}[(\hat x - x - \mu_{\epsilon})^2] + \text{Var}_{\epsilon}$, and the minimizer of the data loss becomes $\hat x = x + \mu_{\epsilon}$, where $\mu_{\epsilon}$ denotes the mean of the noise distribution.
For the PDE loss, on the other hand, the minimizer remains $\hat x = x$ (assuming that the correct parameters $\lambda$ of $\mathcal{F}(\lambda)$ are known).

The two losses are now counteracting each other and the optimization procedure will produce some compromise between them, even when the parameters $\lambda$ of the PDE are known exactly a priori. 
In the case where the parameters of $\mathcal{F}(\lambda)$ are unknown, the optimization procedure will tend to converge to parameter values which result in a lower least-squares loss than the correct ones might.

\subsection{Reconciling the Losses}

A simple way of restoring consistency between the two losses in the limit of infinite data is to add an offset parameter $\theta_0$ to the PINN prediction $\hat x$ in the data loss term, which is supposed to learn the bias in the noise term.
The data loss then reads $\underset{N_d \rightarrow \infty}{\lim}\mathcal{L}_{\rm data} = \mathbb{E}[(\hat x + \theta_0 - x - \epsilon)^2]= \mathbb{E}[(\hat x  + \theta_0 - x - \mu_{\epsilon})^2] + \text{Var}_{\epsilon}$ and the optimum $\hat x = x$, $\theta_0 = \mu_{\epsilon}$ would be compatible with the PDE loss.
Hence the optimization procedure could converge to the correct solution, in principle.

However, since the maximum likelihood estimator is asymptotically optimal only when using the correct likelihood \citep{Wasserman2004}, this data loss still does not constitute the best option (except for the case of Gaussian noise with non-zero mean).
Furthermore, in the practical case of finite data, outliers may have an outsized effect when using the least-squares loss.
Hence, a way of taking the non-Gaussianity of the noise into account explicitly would likely further improve the speed of optimization as well as the final learning outcome.
In the next section, we demonstrate how EBMs can be used for this purpose.

\subsection{Using EBMs to Learn the Noise Distribution}
In order to also take the shape of non-Gaussian noise into account, we can choose to use a different data loss in \eqref{eq:lossges}.
A natural choice would be the log-likelihood of the measurements, given the noise distribution.
In case of Gaussian noise, this would again result in a least-squares loss term, plus a constant.
However, since in our problem setup we do not know the form of the noise a priori, we need to determine the shape of the noise distribution jointly with the PINN solution.

\begin{algorithm2e}[t]
\SetKwInOut{Input}{Input}
\SetKwInOut{Output}{Output}
\caption{Training the PINN-EBM}
\Input{$N_{\rm tot}$; $N_{\rm EBM}$; PINN $\hat x (\cdot| \theta_{\rm PINN})$, EBM $\hat h(\cdot| \theta_{\rm EBM})$, data points $\mathcal{D}_d$,\\
collocation points $\mathcal{D}_c$, differential operator $\mathcal{F}(\lambda)$, weighting factor $\omega$}
\Output{optimized $\theta_{\rm PINN}$, $\theta_{\rm EBM}$, $\lambda$}
$i=0$ \\
\While{$i \leq N_{\rm tot}$}
{
    Draw a minibatch of data points $\{t_d,y_d\}_{\rm mb}$ and of collocation points $\{t_c\}_{\rm mb}$ \\ 
    \uIf{$i < i_{\rm EBM}$}
    {
        Calculate data loss $\mathcal{L}_{\rm data}(\hat x, \{t_d, y_d\}_{\rm mb})$ according to~\eqref{eq:data_loss}
    }
    \uElse{
        \lIf{$i = i_{\rm EBM}$}{initialize EBM for $N_{\rm EBM}$ iterations \DontPrintSemicolon}
        Calculate data loss $\mathcal{L}_{\rm data}^{\rm EBM}\left(\{y_d - \hat x(t_d)\}_{\rm mb}, \hat h\right)$ from \eqref{eq:lossges_ebm}
    }
    \textbf{end} \\
    \leIf{$i >= i_{\rm EBM}$}{$\omega' = \omega$ \DontPrintSemicolon}{$\omega' = 1$}
    Calculate PDE loss $\mathcal{L}_{\rm PDE}(\mathcal{F}, \hat x, \{t_c\}_{\rm mb})$ according to \eqref{eq:pde_loss} \\ 
    Compute total loss $\mathcal{L}_{\rm tot} = \mathcal{L}_{\rm data} + \omega' \mathcal{L}_{\rm PDE}$ \\
    Calculate gradient $\nabla_{\theta} \mathcal{L}_{\rm tot}$ and update $\theta = \{ \theta_{\rm PINN}, \theta_{\rm EBM}, \lambda\}$ \\
    $i \mathrel{+}= 1$
}
\label{alg:pinn_ebm_training}
\end{algorithm2e}

To this end, we train an EBM and employ it to obtain the negative log-likelihood of the data, given both our models.
For training, we then utilize the following loss function:
\begin{equation} \label{eq:lossges_ebm}
    \mathcal{L}_{\rm tot} = \mathcal{L}_{\rm data}^{\rm EBM}(\{y_d-\hat x(t_d)\}_{\rm mb}, \hat h| \theta_{\rm PINN}, \theta_{\rm EBM}) + \omega \mathcal{L}_{\rm PDE}(\mathcal{F}, \hat x, \{t_c\}_{\rm mb}| \theta_{\rm PINN}, \lambda),
\end{equation}
where $\mathcal{L}_{\rm data}^{\rm EBM}(\cdot, \hat h) = \frac{1}{N_d'} \text{NLL}(\cdot, \hat h) = \log{Z(\hat h)} - \frac{1}{N_d'}\sum_{i=1}^{N_d'} \hat h(\cdot)$, utilizing \eqref{eq:NLL}; note that $\hat h$ and hence the NLL are now independent of $t$, because we consider homogeneous noise.
Since we do not know the actual magnitude of the noise at each data point, we use the residuals $y_d-\hat x(t_d)$ between the current PINN prediction and the measurements as our best guess. 
These estimates of the noise values then serve as training data for the EBM.
In other words, we employ an unconditional EBM to model the PDF of the residuals.
With the loss \eqref{eq:lossges_ebm}, both models are trained jointly until convergence.

\subsection{Training PINN and EBM Jointly}

In Algorithm \ref{alg:pinn_ebm_training}, the training procedure is summarized.
Since both PINN and EBM need to be trained in parallel, the optimization procedure can be very challenging.
In our experiments, it proved advantageous to start with the standard PINN loss in order to obtain a solution close to the data, and to only subsequently, after $i_{\rm EBM}$ iterations, switch to the EBM loss, in order to fine-tune the solution.
Before the EBM loss is used for the first time, we initialize the EBM by training it for $N_{\rm EBM}$ iterations on the current noise estimates $y_d - \hat x(t_d)$, while keeping all parameters except $\theta_{\rm EBM}$ fixed.
When we choose to use a weighting parameter $\omega \neq 1$, we also wait until the EBM has been initialized before switching to the new value.
In total, we train for $N_{\rm tot}$ iterations.

Initializing the EBM only later during the training process is advantageous for two reasons:
firstly, the least-squares loss is still a sensible first guess, so initializing the EBM on the residuals stemming from the pretrained network will let it start out with a reasonable form of the likelihood instead of a random one.
Secondly, it will give us a better idea about the range in which the residuals between the PINN prediction and the data lie, allowing for a better normalization of the inputs to the EBM and hence more efficient training.

\section{Experiments \protect\footnote{The code for this project is available at \href{https://github.com/ppilar/PINN-EBM}{https://github.com/ppilar/PINN-EBM}.}} \label{sec:experiments}

In our experiments, we compare the performance of the standard PINN from Section \ref{sec:PINN} to that of the PINN with learnable offset parameter $\theta_0$ and the combination of PINN and EBM.
We will refer to these models as PINN, PINN-off and PINN-EBM.

In our experiments, we consider homogeneous noise in a variety of shapes: Gaussian noise $p(\epsilon) = \mathcal{N}(\epsilon|0,2.5^2)$, a uniform distribution $p(\epsilon) = \mathcal{U}[0,10]$ and a Gaussian mixture of the form $p(\epsilon) = \frac{1}{3}\left(\mathcal{N}(\epsilon|0,2^2) + \mathcal{N}(\epsilon|4,4^2) + \mathcal{N}(\epsilon|8,0.5^2)\right)$.
We also consider the same Gaussian mixture, shifted to have zero mean.
We refer to these noise distributions as G, u, 3G, and 3G0.
The values given here are scaled to the example of the exponential function in Section \ref{sec:exp}.
For other experiments, we use rescaled versions of the same distributions.

When evaluating the performance of our models, the following metrics are considered:
the absolute error in the learned values of the PDE parameters ($|\Delta \lambda|$), the root-mean-square error (RMSE) between $\hat x$ and $x$ in the range of the validation data, the negative log-likelihood (NLL) of the validation data according to the models, and the square values of the PDE residuals ($f^2$) in the range of the training data.
In practice, the NLL is the most relevant performance metric: contrary to the RMSE and $|\Delta \lambda|$, it can also be calculated when the true solution is not known.
While $f^2$ can also be calculated without knowledge of the true solution, it only measures how well the learned PDE is fulfilled and not necessarily the correct one.

\ifthenelse
{\boolean{arxiv}}
{Additional results are discussed in Appendices \ref{app:experiments} and \ref{app:parameters}:}
{Additional results are available in the extended report on arXiv \citep{pilar2022physics}:}
the method is tested on another PDE, and the model performance is investigated as a function of various dataset properties and training parameters.

\subsection{Toy Problem} \label{sec:exp}

\begin{wraptable}{r}{5cm}
\vspace{-0.85cm}
\includegraphics[width=\linewidth]{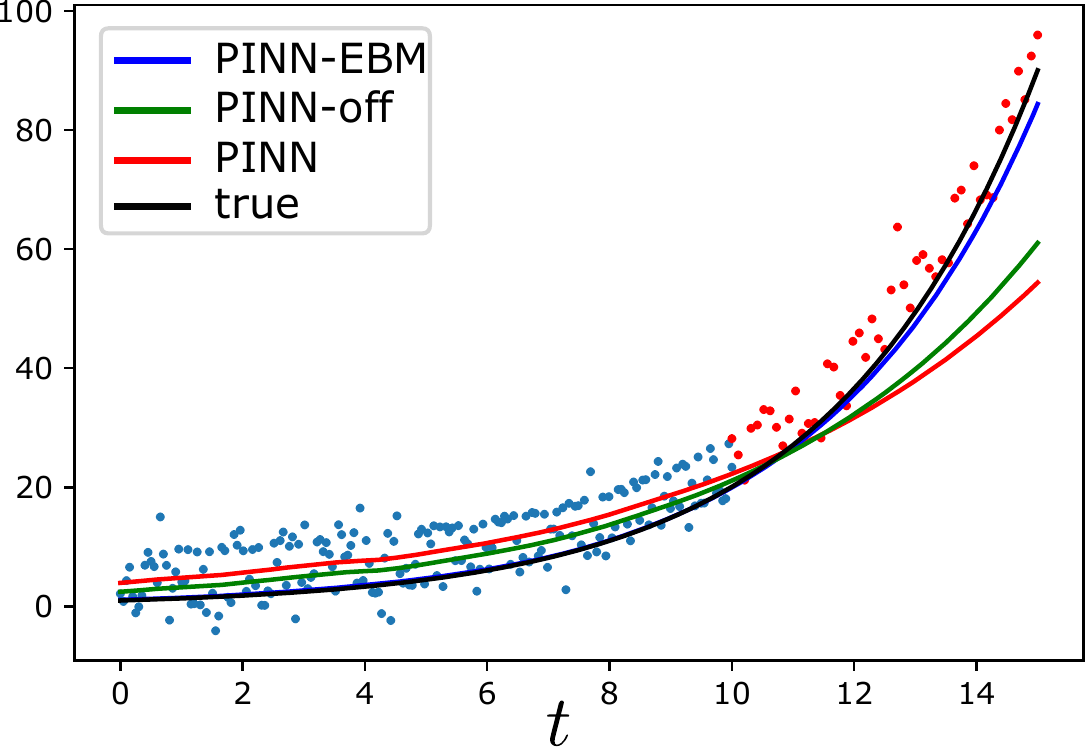}
    \captionof{figure}{The exponential differential equation \eqref{eq:exp_de} with Gaussian mixture noise (3G).
    Results for one individual run, where the blue dots represent training data and the red dots validation data.}
	\label{fig:exp_example}
\end{wraptable}
As a first toy problem, we consider the following simple differential equation,
\begin{equation} \label{eq:exp_de}
    \dot x(t) - \lambda x(t) = 0, 
\end{equation}
where $\lambda = 0.3$ and with the true solution being the exponential function, $x(t) = e^{\lambda t}$.

\begin{wraptable}{r}{9.5cm}
    \centering
    \begin{tabular}{llrclll}
        \toprule        
        \multicolumn{2}{l}{noise} & PINN-EBM & PINN-off & PINN \\        
        \cmidrule{3-5}
        \cmidrule{1-1}
        3G
& $100|\Delta \lambda|$  & \textbf{1.2}$\pm$1.1 & 4.0$\pm$3.6 & 10.2$\pm$1.5\\
& RMSE & \textbf{1.9}$\pm$1.4 & 6.1$\pm$6.8 & 8.3$\pm$1.9 \\
& NLL & \textbf{3.5}$\pm$1.0 & 7.9$\pm$9.0 & 8.9$\pm$2.8 \\
& $100f^2$ & \textbf{0.5}$\pm$0.3 & 30.9$\pm$11.5 & 48.8$\pm$17.1  \\

        \cmidrule(lr){3-5} 
        u
& $100|\Delta \lambda|$  & \textbf{1.3}$\pm$0.3 & 3.3$\pm$2.5 & 12.3$\pm$0.7 \\
& RMSE & \textbf{1.9}$\pm$0.8 & 4.0$\pm$3.7 & 9.1$\pm$1.1 \\
& NLL & \textbf{4.8}$\pm$1.1 & 6.5$\pm$6.8 & 19.1$\pm$3.8  \\
& $100f^2$ & \textbf{0.7}$\pm$0.3 & 16.3$\pm$4.8 & 37.9$\pm$12.6  \\
        
        \cmidrule(lr){3-5} 
        G
& $100|\Delta \lambda|$  & 2.5$\pm$1.8 & 2.0$\pm$1.3 & \textbf{1.1}$\pm$0.9 \\
& RMSE & 3.4$\pm$2.1 & 3.2$\pm$2.7 & \textbf{2.8}$\pm$2.5 \\
& NLL & 5.1$\pm$2.5 & 5.1$\pm$3.6 & \textbf{4.2}$\pm$2.5  \\
& $100f^2$ & \textbf{0.4}$\pm$0.3 & 10.8$\pm$6.0 & 11.7$\pm$6.6 \\

        \cmidrule(lr){3-5}         
        3G0
& $100|\Delta \lambda|$  & \textbf{1.2}$\pm$1.0 & 5.3$\pm$3.2 & 2.2$\pm$2.3  \\
& RMSE & \textbf{2.2}$\pm$1.4 & 7.9$\pm$6.9 & 5.0$\pm$5.1  \\
& NLL & \textbf{3.7}$\pm$1.3 & 9.8$\pm$11.2 & 5.4$\pm$4.0  \\
& $100f^2$ & \textbf{0.6}$\pm$0.4 & 32.2$\pm$15.1 & 34.0$\pm$15.5  \\

        \bottomrule \\
    \end{tabular}
    \caption{
    Results for the exponential differential equation~\eqref{eq:exp_de}. 
    The entries in the table are averages obtained over 10 runs plus-or-minus one standard deviation. Bold font highlights best performance.
    }
    \label{tab:exp}
\end{wraptable}

In Fig. \ref{fig:exp_example}, the results for an individual run of PINN, PINN-off, and PINN-EBM are compared;
blue dots represent training data and red dots validation data.
From this plot, it is clear that PINN converges to a wrong solution, and PINN-off gives only a slightly better prediction, whereas the PINN-EBM result is very close to the true curve.

In Table \ref{tab:exp}, performance metrics of the different models as obtained for different noise forms and with $\omega=1$ are given.
The results have been averaged over ten runs with different realizations of the measurement noise.

We start by discussing the results for Gaussian mixture noise (3G).
It is evident that PINN performs poorly on all metrics.
While PINN-off learns the correct parameter $\lambda$ better on average, the variance of its predictions is very high, thereby limiting its use.
The PINN-EBM estimate of the parameter is significantly less variable than that of PINN-off, and is very close to the correct value of $\lambda$.
In accordance with the quality of the parameter estimates, PINN-EBM clearly outperforms both PINN and PINN-off when considering the RMSE and NLL; it achieves the lowest RMSE as well as the lowest NLL on the validation data.

When considering $f^2$, PINN needs to trade off accuracy in this metric with minimizing the data loss \eqref{eq:data_loss}, due to the inconsistency between the losses discussed in Section \ref{sec:PINN_nonzero}.

Three other noise forms are considered in the table:
for the uniform (u) noise distribution, the results are very similar to the Gaussian mixture.
In case of Gaussian (G) noise, PINN performs best, which makes sense since it is implicitly using the correct likelihood.
PINN-EBM, on the other hand, performs a bit worse as there is some overfit to the noise in the data.

For Gaussian mixture noise with zero mean (3G0), where the PDE and data loss are no longer misaligned for the PINN, PINN-EBM still outperforms PINN, although with a lesser margin.
This shows that PINN-EBM can successfully mitigate the effects of non-Gaussian noise and significantly improves the quality of the parameter estimates, as well as that of the solution to the regression task.
While PINN-off manages to improve upon PINN in the case of non-zero mean noise, the high variance in its predictions makes it less reliable than PINN-EBM.

\begin{figure}[t]
\centering
	\includegraphics[width=\linewidth]{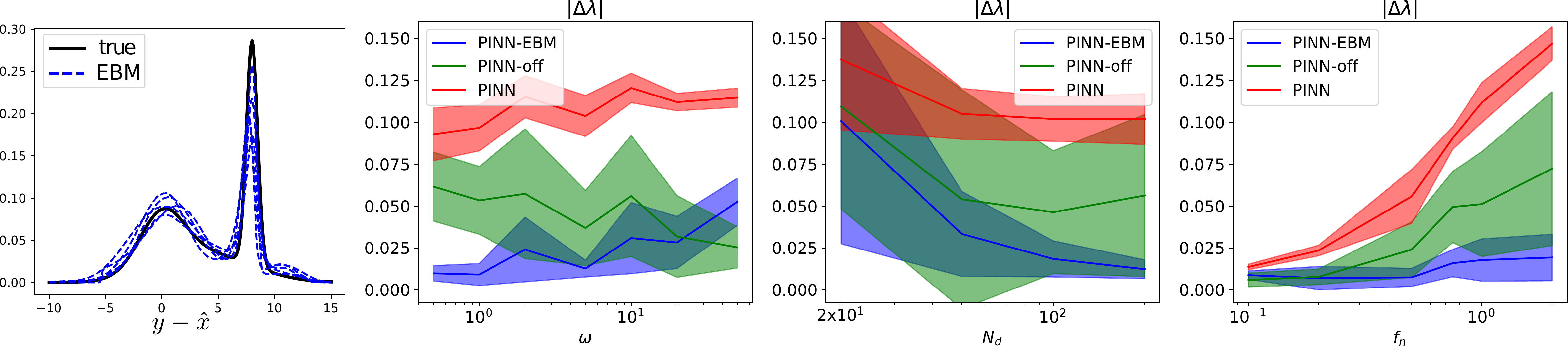}
	\caption{\textbf{Left:} The Gaussian mixture (3G) noise distribution, together with examples of the PDFs learned by the EBM. \textbf{Middle left:} Performance of the models as a function of the weighting factor~$\omega$. \textbf{Middle right:} Performance of the models as a function of the number of training points $N_d$. \textbf{Right:} Performance of the models as a function of the noise strength (the distribution in the left plot is scaled by the factor $f_n$). The standard values are $\omega = 1$, $N_d=200$, and $f_n=1$. The averages of 5 runs are depicted, plus-or-minus one standard deviation.}
	\label{fig:exp_results_summary}
\end{figure}

In Fig. \ref{fig:exp_results_summary}, the impact of various training parameters and dataset properties on the model performance is shown, when considering 3G noise (left plot).
It is apparent that the weighting parameter~$\omega$ can significantly impact the model performance. 
From the two plots on the right, we observe that the performances of the three models converge as either the number of training points or the noise strength is reduced;
for higher values, the impact of non-Gaussian noise is larger and PINN-EBM outperforms.
\ifthenelse
{\boolean{arxiv}}
{More in-depth discussions of these results can be found in Appendix \ref{app:parameters}.}
{More in-depth discussions of these results can be found in
the appendices of \citet{pilar2022physics}.}

\subsection{Navier-Stokes Equations} \label{sec:NS}

For our next experiment, we consider the example of incompressible flow past a circular cylinder and use the dataset from \citet{Raissi2019}, to which we add Gaussian mixture (3G) noise.
This setup can be described by the 2D Navier-Stokes equations,
\begin{subequations} \label{eq:fg}
\begin{align} 
    u_{\rm t} + \lambda_1(u u_{\rm x} + vu_{\rm y}) + p_{\rm x} - \lambda_2 (u_{\rm xx} + u_{\rm yy}) &= 0, \\
    v_{\rm t} + \lambda_1(uv_{\rm x} + vv_{\rm y}) + p_{\rm y} - \lambda_2(v_{\rm xx} + v_{\rm yy}) &= 0,
\end{align}
\end{subequations}
where $u$ and $v$ denote the components of the velocity field and $p$ the pressure; the subscripts indicate derivatives with respect to the different dimensions of the inputs $t = (\mathrm{t, x, y})$. Here, the PINN does not output $u$ and $v$ directly, but instead models the potential $\psi$ and the pressure $p$; then $u=\psi_{\rm y}$ and $v=-\psi_{\rm x}$. The true values of the parameters are $\lambda_1=1$ and $\lambda_2=0.01$.

The results are given in Table \ref{tab:NS_data}.
It is apparent that PINN-EBM learns $\lambda_1$ better than PINN and PINN-off whereas the difference is smaller for $\lambda_2$.
PINN-EBM also performs better in terms of RMSE and NLL.
Here, we have chosen $\omega = 50$ for PINN-EBM since the NLL on the validation data decreased when the PDE loss was weighted more strongly (see Fig.~\ref{fig:NS_omega}).
For PINN and PINN-off, increasing the parameter did not lead to notable changes in the NLL and we kept $\omega=1$.

\subsection{Implementation Details} \label{sec:implementation}

\begin{table}
\begin{minipage}{0.3\linewidth}
\centering
	\includegraphics[width=\linewidth]{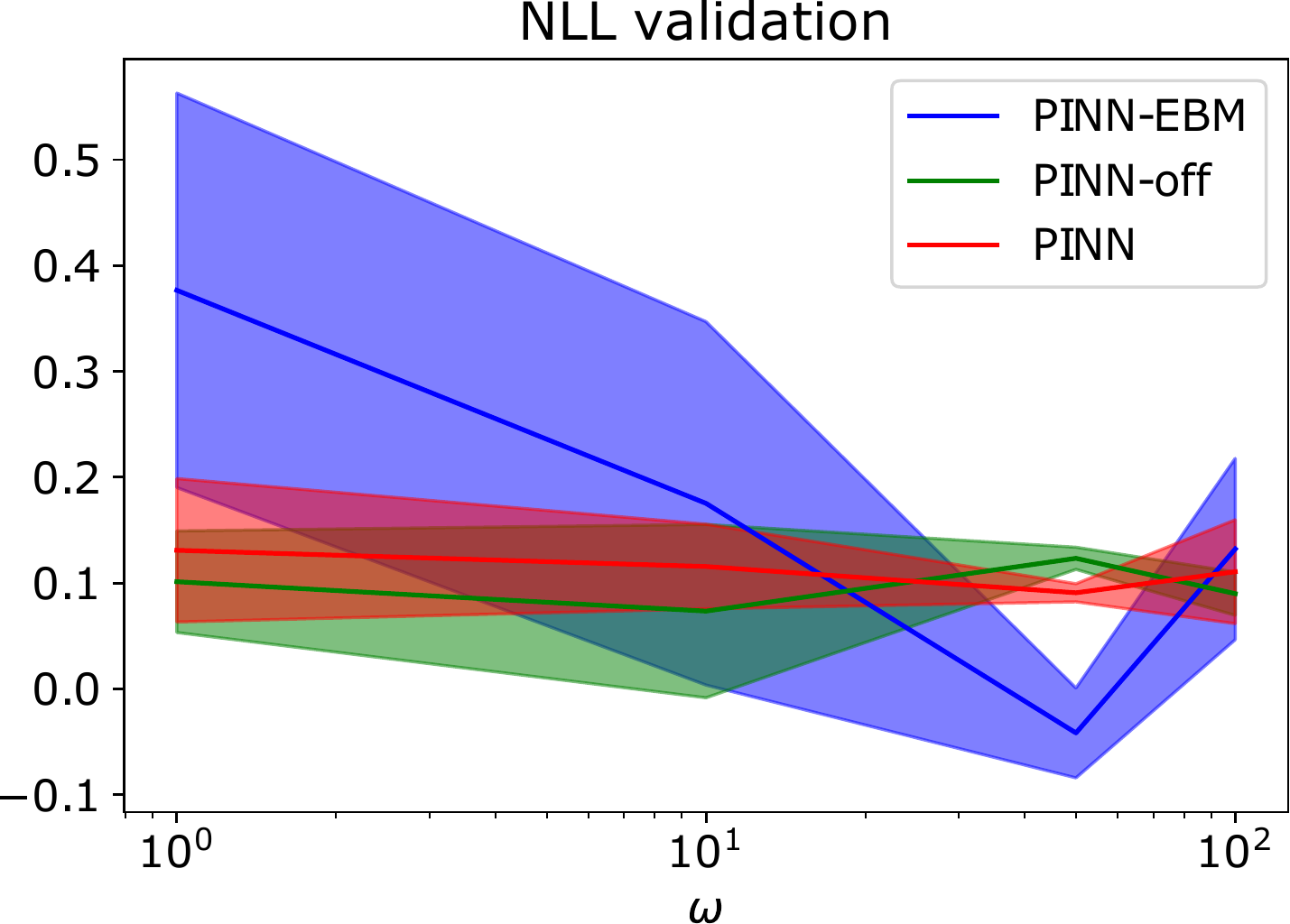}
    \captionof{figure}{
    The NLL on the validation data as a function of the parameter $\omega$, weighting the PDE loss.
    The averages of 5 runs are depicted, plus-or-minus one standard deviation.
    }
	\label{fig:NS_omega}
\end{minipage}\hfill
\begin{minipage}{0.65\linewidth}
\centering
    \begin{tabular}{llrclc}
        \toprule        
        \multicolumn{2}{l}{noise} & PINN-EBM & PINN-off & PINN \\        
        \cmidrule{3-5}
        \cmidrule{1-1}
        3G
& $100|\Delta \lambda_1|$  & \textbf{1.19}$\pm$0.67 & 2.92$\pm$0.47 & 23.10$\pm$0.11  \\
& $100|\Delta \lambda_2|$  & \textbf{0.04}$\pm$0.03 & 0.09$\pm$0.05 & 0.08$\pm$0.06  \\
& RMSE & \textbf{0.06}$\pm$0.01 & 0.11$\pm$0.01 & 0.20$\pm$0.02 \\
& NLL & \textbf{-0.03}$\pm$0.08 & 0.15$\pm$0.07 & 0.40$\pm$0.30  \\
& $100f^2$ & \textbf{0.04}$\pm$0.00 & 0.10$\pm$0.01 & 0.18$\pm$0.01  \\

        \bottomrule \\
    \end{tabular}
    \caption{
    Results for the Navier Stokes equations \eqref{eq:fg} in case of Gaussian mixture (3G) noise.    
    The entries in the table are averages obtained over 5 runs plus-or-minus one standard deviation.    
    Bold font highlights best performance.
    }
\label{tab:NS_data}
\end{minipage}	
\end{table}

For both PINN and EBM, fully-connected neural networks with tanh activation function were used. 
Both inputs and outputs of the networks were normalized to their expected ranges.
For the PINN, 4 layers of width 40 were chosen for the exponential equation;
in case of the Navier-Stokes equations, 8 layers of width 20 were used instead.
For the EBM, 3 layers of width 5 were used and a dropout layer with factor $0.5$ was inserted before the last layer.

The PINN was trained for $40.000$ iterations in case of the exponential equation (Table \ref{tab:exp}) and $100.000$ iterations for the Navier-Stokes equations (Table \ref{tab:NS_data}).
When evaluating the impact of training parameters (Figs. \ref{fig:exp_results_summary}, \ref{fig:NS_omega}), $50.000$ iterations were used for both of them.
The EBM was initialized for $N_{\rm EBM} = 2000$ iterations, after $i_{\rm EBM}=4000$ and $i_{\rm EBM}=10000$ iterations, respectively, for the exponential equation and the Navier-Stokes equations.
Convergence was checked by confirming that the values of the PDF tended toward zero on both ends of the distribution;
if the check failed, the initialization was repeated with an increased number of iterations until it passed.

The datasets employed in Section \ref{sec:exp} contained 200 training points, 50 validation points and 2000 collocation points.
For the Navier-Stokes example in Section \ref{sec:NS}, 4000 training points, 1000 validation points and 4000 collocation points were used.
The collocation points were generated on a uniform grid.
The Adam optimizer with a learning rate of $0.002$ was used for both PINN and EBM.
In case of the Navier-Stokes equations, the learning rate was reduced by a factor of $0.3$ after $80\,000$ iterations.
The batch size for data points was 200 (i.e. full batches were used for the exponential equation) and the batch size for collocation points was 100.
Training the PINN-EBM typically took 30-50\% longer than the standard PINN.

\section{Conclusions and Future Work} \label{sec:conclusions}
In this paper, we demonstrated that the standard PINN fails in case of non-zero mean measurement noise and we proposed the PINN-EBM to resolve this problem;
utilizing an EBM to learn the homogeneous noise distribution allows the PINN to produce good results also in case of non-zero mean and non-Gaussian noise.
Using several examples, ranging from a simple toy problem to the complex Navier-Stokes equations, we demonstrated the capabilities of our method and showed that it outperforms the standard PINN by a significant margin in case of non-Gaussian noise.
In addition to determining the correct PDE solution, the PINN-EBM also allows for the identification of the true noise distribution, which may result in novel insights into the measurement procedure.

In the future, it would be interesting to investigate combining the PINN-EBM with other improvements to the PINN framework, such as adaptive weighting schemes 
\citep{McClenny2020, xiang2022self}, adaptive collocation point placement \citep{wu2023comprehensive}, scheduling approaches \citep{Krishnapriyan2021, wang2024respecting} or additional loss terms \citep{Yu2022}.
Extending the approach to certain kinds of heteroscedastic noise and investigating the applicability of the approach to other frameworks such as neural ODEs \citep{chen2018neural} constitute promising avenues for future research.
Finally, it would be interesting to apply the method to suitable real-world datasets.

\section*{Acknowledgements}

The work is financially supported by the Swedish Research Council (VR) via the project \emph{Physics-informed machine learning} (registration number: 2021-04321) and by the \emph{Kjell och M{\"a}rta Beijer Foundation}. We would also like to thank Fredrik K. Gustafsson for valuable feedback.

\bibliography{pebm}

\newpage

\appendix

\section{Details on the Experiments} \label{app:experiments}

In this appendix, we present one additional experiment and give more details on the experiments described in Section \ref{sec:experiments}.

\subsection{Additional Experiment: the Bessel Equation} \label{sec:Bessel}

To test the framework on a more complicated differential equation than the exponential equation \eqref{eq:exp_de}, we employ the Bessel equation, a second-order differential equation \citep{Niedziela2008, Bowman2012}:
\begin{equation} \label{eq:Bessel}
(\lambda t)^2 \ddot x(t) + \lambda t \dot x(t) + ((\lambda t)^2 - \nu^2) x(t) = 0,
\end{equation}
where we pick $\nu = 1$ and where we have introduced the parameter $\lambda =0.7$, which is to be estimated by the PINN. 
Results for one run of this experiment are shown in Fig. \ref{fig:Bessel_example} and results for multiple runs are given in Table \ref{tab:Bessel}.
When comparing the results obtained here to those from Section \ref{sec:exp}, we note the following:
PINN-EBM outperforms both PINN and PINN-off on all metrics except $f^2$, and PINN-off performs better than PINN.
For this example, the variance of the results for both PINN and PINN-off is smaller than for the exponential equation.
One reason for this can be found in the $t^2$ term in~\eqref{eq:Bessel}, which can lead to large values in the PDE loss in \eqref{eq:lossges}, thereby putting more weight on PDE fulfillment than on matching the data to a Gaussian.

Another reason why PINN and PINN-off perform better here than for the toy example lies in the shape of the PDE solution:
in the previous example from Section \ref{sec:exp}, rather inaccurate estimates of the parameter $\lambda$ could still have been compatible with Gaussian noise of variable strength.
Here, on the other hand, the parameter determines the frequency of the oscillation of the curve, leaving less wiggle room for compromise.

\begin{table}
\begin{minipage}{0.3\linewidth}
\centering
	\includegraphics[width=\linewidth]{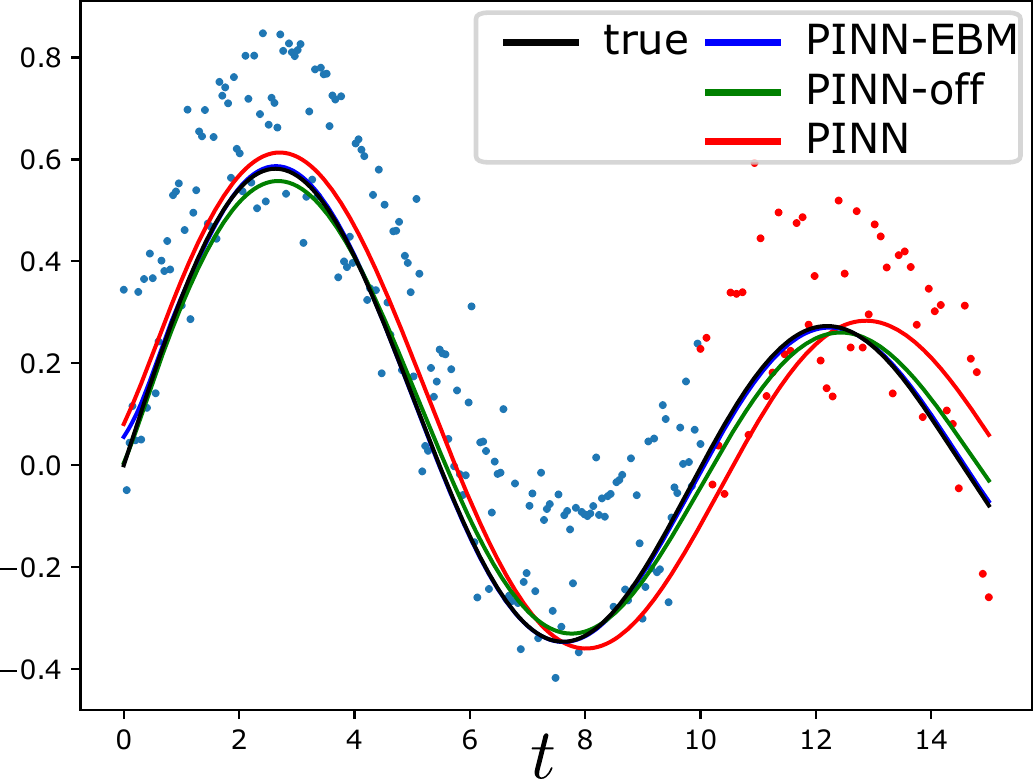}
    \captionof{figure}{
    The Bessel equation \eqref{eq:Bessel} with Gaussian mixture noise (3G).
    Results for one individual run, where the blue dots represent training data and the red dots validation data.
    }
	\label{fig:Bessel_example}
\end{minipage}\hfill
\begin{minipage}{0.65\linewidth}
\centering
    \begin{tabular}{llrclc}
        \toprule        
        \multicolumn{2}{l}{noise} & PINN-EBM & PINN-off & PINN \\        
        \cmidrule{3-5}
        \cmidrule{1-1}
        3G
        & $100|\Delta \lambda|$  & \textbf{0.27}$\pm$0.22 & 1.81$\pm$0.93 & 3.11$\pm$1.58  \\
        & RMSE & \textbf{0.00}$\pm$0.00 & 0.04$\pm$0.02 & 0.06$\pm$0.03 \\
        & NLL & \textbf{-0.94}$\pm$0.13 & -0.63$\pm$0.08 & -0.19$\pm$0.32  \\
        & $100f^2$ & 0.63$\pm$0.14 & 0.39$\pm$0.40 & \textbf{0.19}$\pm$0.05  \\

        \bottomrule \\
    \end{tabular}
    \caption{
    Results for the Bessel differential equation \eqref{eq:exp_de} in case of Gaussian mixture (3G) noise.
    The entries in the table are averages obtained over 10 runs plus-or-minus one standard deviation. Bold font highlights best performance.
    }
\label{tab:Bessel}
\end{minipage}	
\end{table}

\subsection{Noise Forms} \label{app:noise}

\begin{figure}[t]
\centering
	\includegraphics[width=0.9\linewidth]{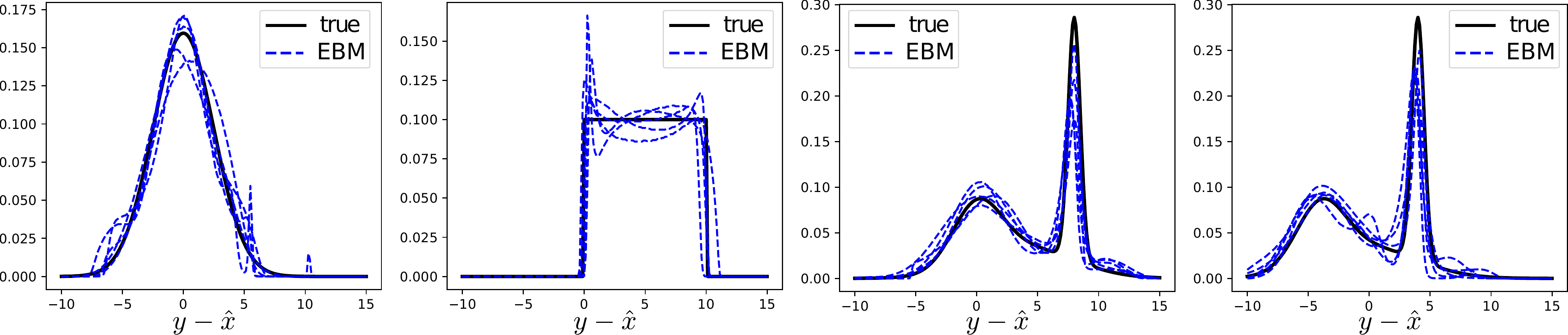}
	\caption{The noise distributions considered in the experiments are depicted. From left to right, we have a Gaussian distribution (G), a uniform distribution (u), a Gaussian mixture (3G), and the same Gaussian mixture, shifted to obtain zero mean (3G0).
    The dashed blue curves give examples of PDFs learned by the EBM during the training process, whereas the black curve gives the true PDF.}
	\label{fig:noise_plot}
\end{figure}

In Fig. \ref{fig:noise_plot}, plots of the noise distributions defined in Section \ref{sec:experiments} are shown, together with examples of the PDFs learned by the EBM.
Overall, the EBM approximations are in good agreement with the true noise distributions.

\subsection{Learning Curves} \label{app:lcurves}

\begin{figure*}[!htb]
\centering
	\includegraphics[width=\linewidth]{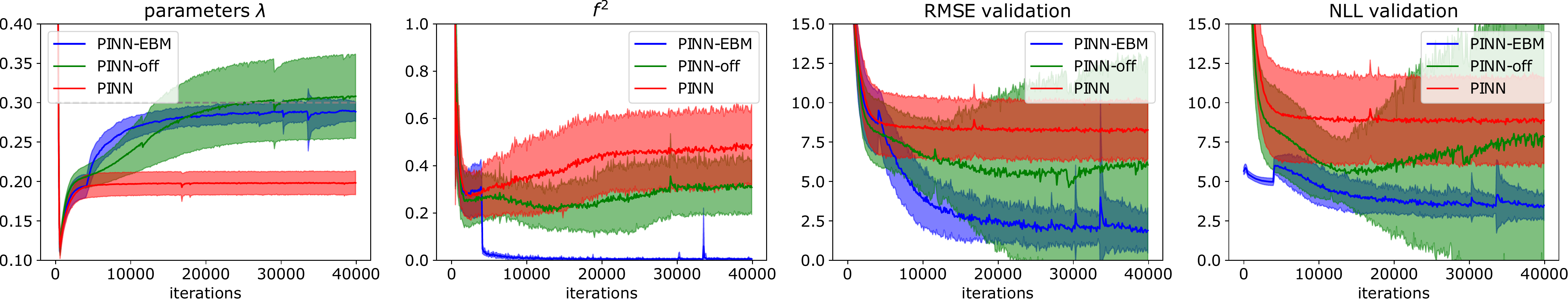}
	\caption{ 
    Learning curves for the exponential differential equation (Section \ref{sec:exp}) with 3G noise.
    The averages of 10 runs are depicted, plus-or-minus one standard deviation.
    }
	\label{fig:exp_lcurves}
\end{figure*}

\begin{figure*}[!htb]
\centering
	\includegraphics[width=\linewidth]{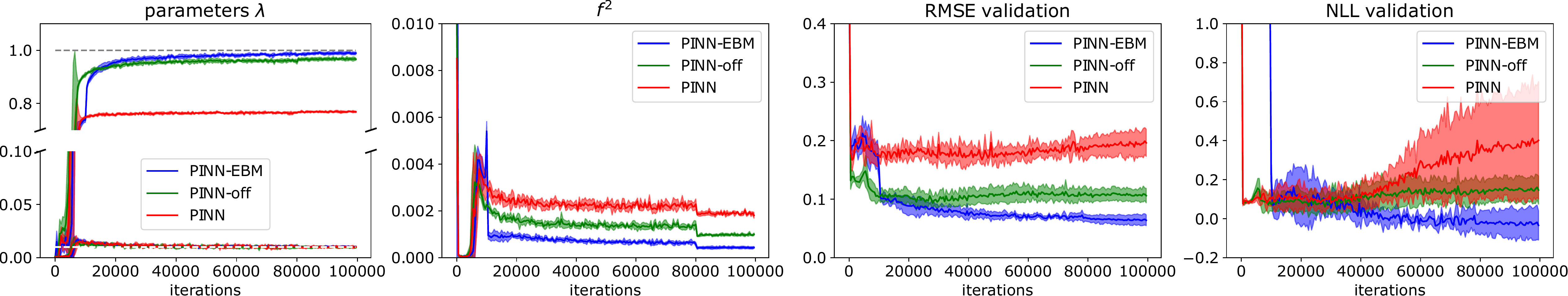}
	\caption{ 
    Learning curves for the Navier-Stokes equations (Section \ref{sec:NS}) with 3G noise.
    The averages of 5 runs are depicted, plus-or-minus one standard deviation.
    }
	\label{fig:NS_lcurves}
\end{figure*}

\begin{figure*}[!htb]
\centering
	\includegraphics[width=\linewidth]{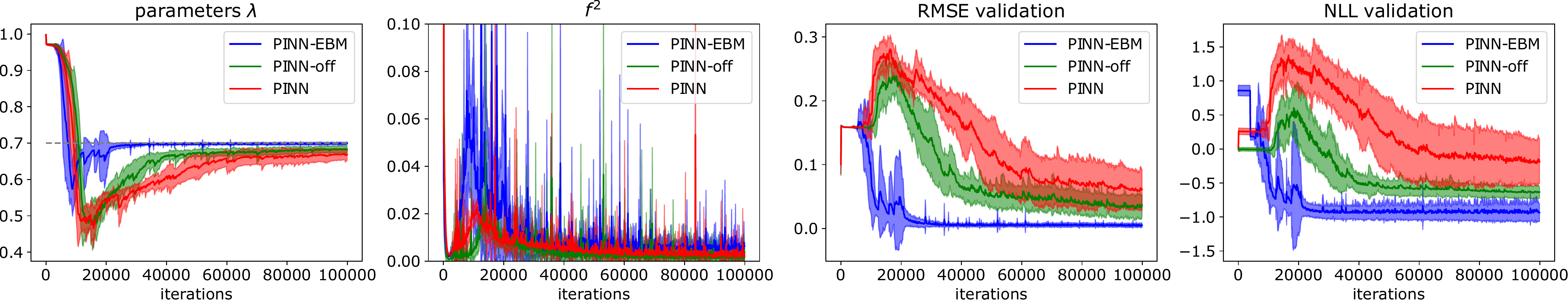}
	\caption{ 
    Learning curves for the Bessel equation (Section \ref{sec:Bessel}) with 3G noise.
    The averages of 10 runs are depicted, plus-or-minus one standard deviation.
    }
	\label{fig:Bessel_lcurves}
\end{figure*}

In Figs. \ref{fig:exp_lcurves}-\ref{fig:Bessel_lcurves}, learning curves for the experiments discussed in Sections \ref{sec:exp}-\ref{sec:NS} and Appendix \ref{sec:Bessel}, with Gaussian mixture (3G) noise, are provided.
From the curves, it becomes apparent that PINN-EBM typically converges quickly to the correct solution after the EBM has been initialized.
In case of the exponential differential equation and the Navier-Stokes equations, PINN and PINN-off converge in a similar amount of iterations, albeit towards often significantly less accurate solutions.
For the Bessel equation, both of them converge towards values close to the correct solution but do so only very slowly.

\subsection{Additional Implementation Details \protect\footnote{The code for this project is available at \href{https://github.com/ppilar/PINN-EBM}{https://github.com/ppilar/PINN-EBM}.}} \label{app:implementation}

For the example of the Bessel equation \ref{sec:Bessel}, the same parameters as for the exponential equation in Section \ref{sec:exp} were used (see Section \ref{sec:implementation});
only the total number of iterations for training the PINN was increased to $100.000$.
For the plots evaluating the impact of training parameters (Figs. \ref{fig:Ntr_ablation}, \ref{fig:fn_ablation}, and \ref{fig:omega_ablations}), the total number of iterations was $50.000$ for all of the experiments.
The noise distributions in Fig. \ref{fig:noise_plot} were scaled by factors of $f_n^0 =1, 0.03$, and $0.05$ for the exponential equation, the Bessel equation, and the Navier-Stokes equations, respectively.

To calculate the NLL for PINN and PINN-off, Gaussian distributions with the following parameters were employed:
the mean was $\mu=0$ in case of PINN and $\mu=\theta_0$ in case of PINN-off.
The standard deviations were estimated from the residuals between the training data and the model predictions.

\section{Impact of Training Parameters} \label{app:parameters}

In this appendix, we study the impact that various training parameters and dataset features have on the model performances when considering the differential equations considered throughout the paper with 3G noise.

\subsection{The Number of Training Points}

\begin{figure*}[t]
\centering
	\includegraphics[width=\linewidth]{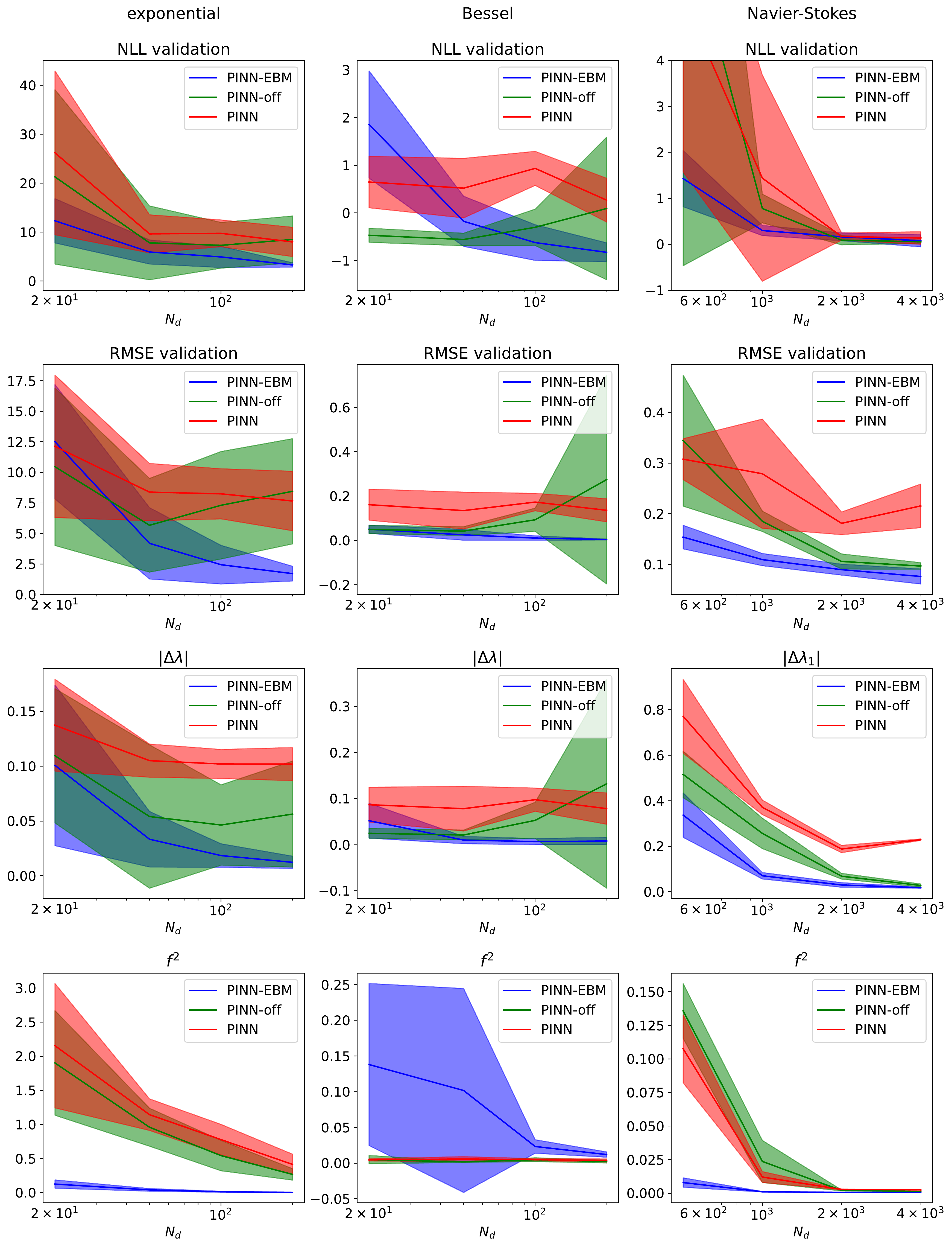}
	\caption{
    The impact of the number of training points $N_d$ on the model performances is depicted for the different experiments considered in Section \ref{sec:experiments} and Appendix \ref{sec:Bessel}, with 3G noise.
    The averages of 5 runs are depicted, plus-or-minus one standard deviation.
 }
	\label{fig:Ntr_ablation}
\end{figure*}

Fig. \ref{fig:Ntr_ablation} depicts the impact of the number of data points available for training. 
The general trend is that the model performances improve with the number of training points.
For PINN-EBM, the trend is most pronounced, which makes sense since more data points will allow it to learn a more accurate noise distribution.
The main exception to this trend is PINN-off in case of the Bessel equation, where the volatility of the predictions increases with the number of training points.
In terms of the accuracy of the PDE solution, as quantified via the RMSE and $|\Delta \lambda|$, the PINN-EBM consistently gives similarly good or better results than both PINN and PINN-off, independent of the number of training points.
This shows that PINN-EBM does not depend on an outsized amount of training data to perform well.

\subsection{The Noise Strength}
\begin{figure*}[t]
\centering
	\includegraphics[width=\linewidth]{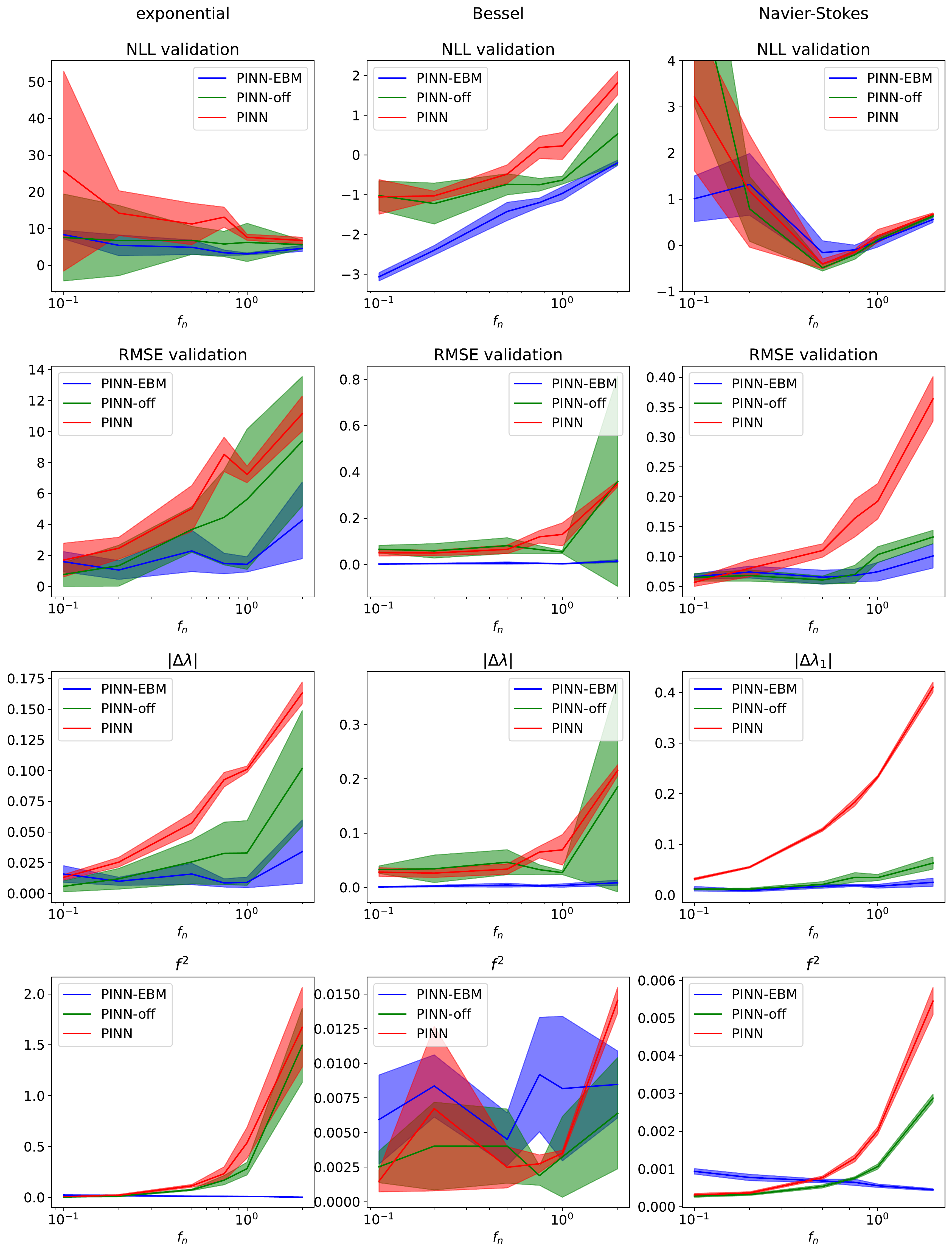}
	\caption{ 
    The impact of the noise strength $f_n$ on the model performances is depicted for the different experiments considered in Section \ref{sec:experiments} and Appendix \ref{sec:Bessel}, with 3G noise.
    The averages of 5 runs are depicted, plus-or-minus one standard deviation.
 }
	\label{fig:fn_ablation}
\end{figure*}

\begin{figure}
\centering
	\includegraphics[width=0.9\linewidth]{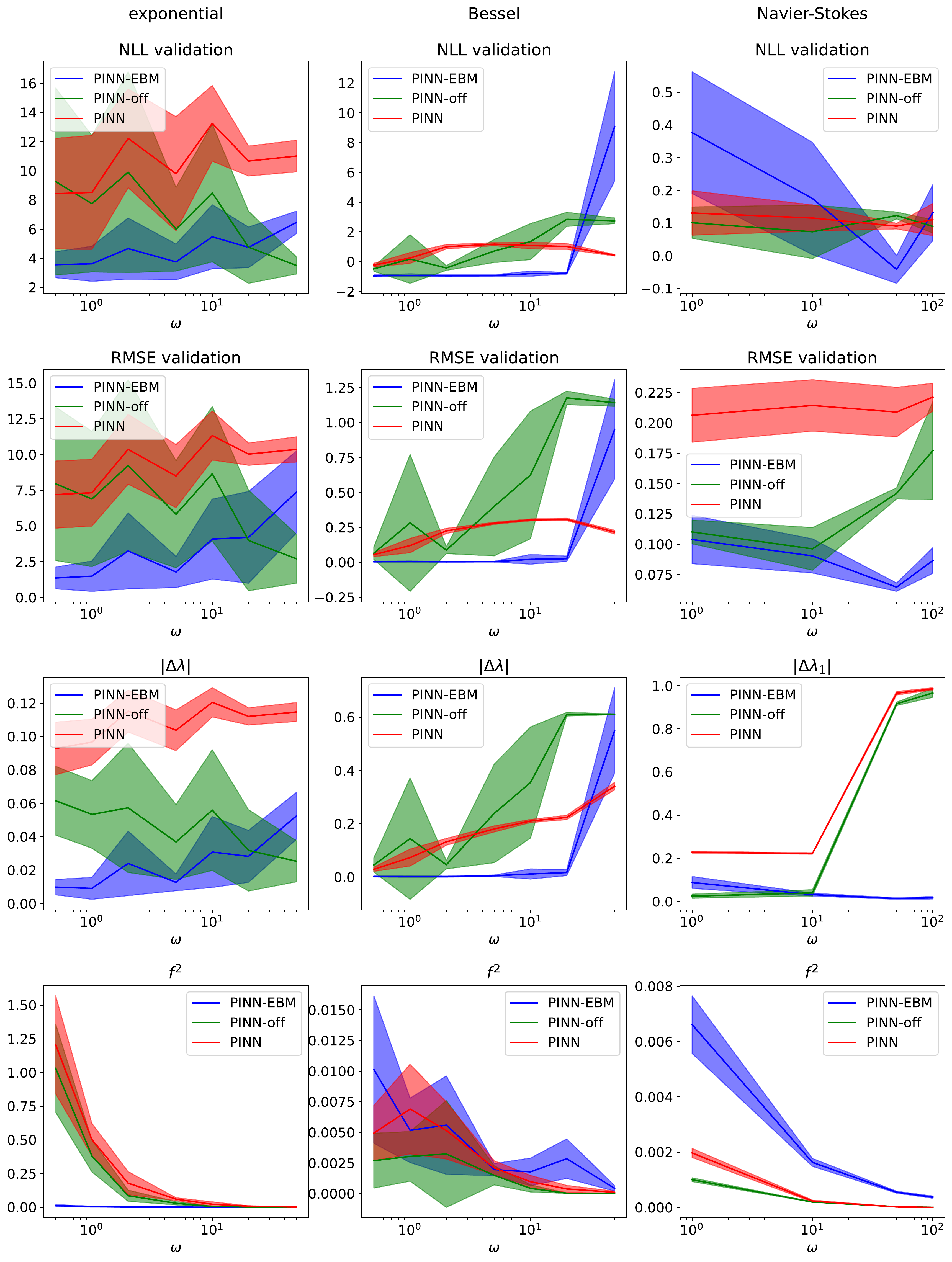}
	\caption{ 
    The impact of the weighting factor $\omega$ on the model performances is depicted for the different experiments considered in Section \ref{sec:experiments} and Appendix \ref{sec:Bessel}, with 3G noise.
    The averages of 5 runs are depicted, plus-or-minus one standard deviation.
    }
	\label{fig:omega_ablations}
\end{figure}

In Fig. \ref{fig:fn_ablation}, the impact of the noise strength on the results is investigated. 
To vary the noise strength, the distributions depicted in Fig.~\ref{fig:noise_plot} are rescaled by a factor of $f_n$ (after rescaling them with the factors $f_n^0$ corresponding to the different experiments, as given in Appendix \ref{app:implementation}).
It is apparent that PINN-EBM clearly outperforms PINN and PINN-off for high noise strengths.
As the noise strength decreases, the performances of the different models converge:
in case of the exponential equation and the Navier Stokes equations, the model performances are roughly the same in case of small noise.
For the Bessel equation, however, the PINN-EBM retains a clear advantage regardless of the noise strength.

\subsection{The Weighting Factor $\omega$ for the PDE Loss} \label{app:omega}

In Fig. \ref{fig:omega_ablations}, results for different values of the weighting factor $\omega$ for the PDE loss in \eqref{eq:lossges} and~\eqref{eq:lossges_ebm} are depicted. 
In the case of PINN-EBM, the training starts with $\omega=1$ and the larger value of $\omega$ is only used as soon as the EBM has been initialized (compare Algorithm \ref{alg:pinn_ebm_training}).
This parameter deserves special attention since it can be tuned in real-world experiments and we are interested in the following question:
is there an optimal value of $\omega$ and how can it be determined?
To address this question, the PDE fulfillment $f^2$ and the NLL are of particular interest since only these quantities are available without knowing the correct PDE solution.

When considering $f^2$, it is apparent that it becomes smaller as $\omega$ increases.
However, the corresponding plots of the RMSE and $|\Delta \lambda|$ show that this does not imply correctness of the results.
In case of the Bessel and the Navier-Stokes equations, the results can become significantly worse as $\omega$ increases.
This also shows that a more strongly weighted PDE loss cannot, in general, override the detrimental effects of the non-Gaussian noise for the standard PINN.
One may have assumed that the PDE would only be compatible with a very limited range of PDE parameters and hence for the PINN to be pushed towards the correct solution that way;
however, this only seems to be the case for PINN-off when applied to the exponential differential equation.

When considering e.g. the plots for the RMSE, it is apparent that no optimal value $\omega$ exists that would be valid for all the differential equations under consideration.
Only PINN seems to consistently perform better for small values of $\omega$.
To determine the best value of $\omega$ for a given case, the NLL on the validation data can be a useful criterion.
For PINN and PINN-off, worse predictions do not always translate into higher values of NLL, as the example of the Navier-Stokes equations shows.
In these cases, it is important to keep an eye on the learned values of $\lambda$ during training, as values $\lambda\approx 0$ can often indicate failure.
For the PINN-EBM, however, there is a clear correlation between lower NLL and better results.
Most importantly, when PINN-EBM achieves a better NLL than the other methods, the corresponding PDE solution is also better.

\end{document}